\documentclass[sigconf]{acmart}
\AtBeginDocument{%
  }

\setcopyright{acmlicensed}
\copyrightyear{2026}
\acmYear{2026}
\acmDOI{XXXXXXX.XXXXXXX}
\acmConference[ICCPS]{17th ACM/IEEE International Conference on Cyber-Physical Systems}{}{Saint Malo, France, May 11-14, 2026}
\acmISBN{978-1-4503-XXXX-X/2025/06}

\usepackage{booktabs,multirow,makecell}
\usepackage[table]{xcolor}
\usepackage{siunitx}
\usepackage{caption}
\PassOptionsToPackage{compatibility=false}{caption}
\usepackage{subcaption}
\usepackage{multirow} 
\usepackage{graphicx}
\usepackage{todonotes}
\usepackage{siunitx}
\usepackage{tikz}

\usetikzlibrary{shapes.geometric, arrows,chains}
\usetikzlibrary{decorations.pathmorphing} 
\usetikzlibrary{decorations.pathreplacing}
\usetikzlibrary{matrix} 
\usetikzlibrary{arrows} 
\usetikzlibrary{calc} 
\usetikzlibrary{svg.path}
\tikzstyle{block} = [draw,rectangle,thick,minimum height=2em,minimum width=2em]
\tikzstyle{sum} = [draw,circle,inner sep=0mm,minimum size=2mm]
\tikzstyle{connector} = [->,thick]
\tikzstyle{line} = [thick]
\tikzstyle{branch} = [circle,inner sep=0pt,minimum size=1mm,fill=black,draw=black]
\tikzstyle{guide} = []
\tikzstyle{snakeline} = [connector, decorate, decoration={pre length=0.2cm,
                         post length=0.2cm, snake, amplitude=.4mm,
                         segment length=2mm},thick, magenta, ->]

\begin{document}

\title{Offline Reinforcement-Learning-Based Power Control for Application-Agnostic Energy Efficiency}

\settopmatter{authorsperrow=4}
\settopmatter{
  printacmref=false,
  printfolios=false
}

\renewcommand\footnotetextcopyrightpermission[1]{}

\author{Akhilesh Raj}
\affiliation{%
  \institution{Vanderbilt University}
  \city{Nashville}
  \country{U.S.A}}
\email{akhilesh.raj@vanderbilt.edu}

\author{Swann Perarnau}
\affiliation{%
  \institution{Argonne National Laboratory}
  \city{Lemont}
  \country{U.S.A}
}

\author{Solomon Bekele Abera}
\affiliation{%
 \institution{Argonne National Laboratory}
 \city{Lemont}
 \country{U.S.A}}

\author{Aniruddha Gokhale}
\affiliation{%
  \institution{Vanderbilt University}
  \city{Nashville}
  \country{U.S.A}}





\begin{abstract}
Energy efficiency has become an integral aspect of modern computing
infrastructure design, impacting the performance, cost, scalability, and durability of production systems. The incorporation of power actuation and sensing capabilities in CPU designs is indicative of this, enabling the deployment of system software that can actively monitor and adjust energy
consumption and performance at runtime.
While reinforcement learning (RL) would seem ideal for the design of such energy efficiency control systems, online training presents challenges ranging from the lack of proper models for setting up an adequate simulated environment, to perturbation (noise) and reliability issues, if training is deployed on a live system.

In this paper we discuss the use of offline reinforcement learning as an alternative approach for the design of an autonomous CPU power controller, with the goal of improving the energy efficiency of parallel applications at runtime without unduly impacting their performance. Offline RL sidesteps the issues incurred by online RL training by leveraging a dataset of state transitions collected from arbitrary policies prior to training. 

Our methodology applies offline RL to a gray-box approach to energy efficiency, combining online application-agnostic performance data (e.g., heartbeats) and hardware performance counters to ensure that the scientific objectives are met with limited performance degradation. 
Evaluating our method on a variety of compute-bound and memory-bound benchmarks
and controlling power on a live system through Intel's Running Average Power Limit, we demonstrate that such an offline-trained agent can  substantially reduce 
energy consumption at a tolerable performance degradation cost.
\end{abstract}




\maketitle

\section{Introduction}\label{sec:Introduction}

As the energy consumption of computing continues to grow, with exascale systems
reaching tens of megawatts and AI datacenters planning for gigawatt
capacity~\cite{chien2023genai}, energy efficiency has become a major
sustainability challenge for the community.
While substantial prior research deals with the design of more energy-efficient hardware or  more efficient algorithms, this paper focuses on improving the energy efficiency of an already deployed system, available to users running arbitrary applications. Energy efficiency, then, is a matter of identifying opportunities for lowering energy consumption (e.g., via power capping) without negatively impacting the performance of applications.

The cornerstone of energy efficiency optimization is the identification of resource imbalance within a system, allowing for a reduction in the power consumption of resources that have minimum impact on performance. Identifying these opportunities requires an approach that is grounded in understanding both the application performance and the impact of hardware power management on that performance. 

\begin{figure}[htb]
\vspace{-0.3cm}
    \centering
    \begin{tikzpicture}
        \node[anchor=south west, inner sep=0] (img)
            {\includegraphics[width=\linewidth]{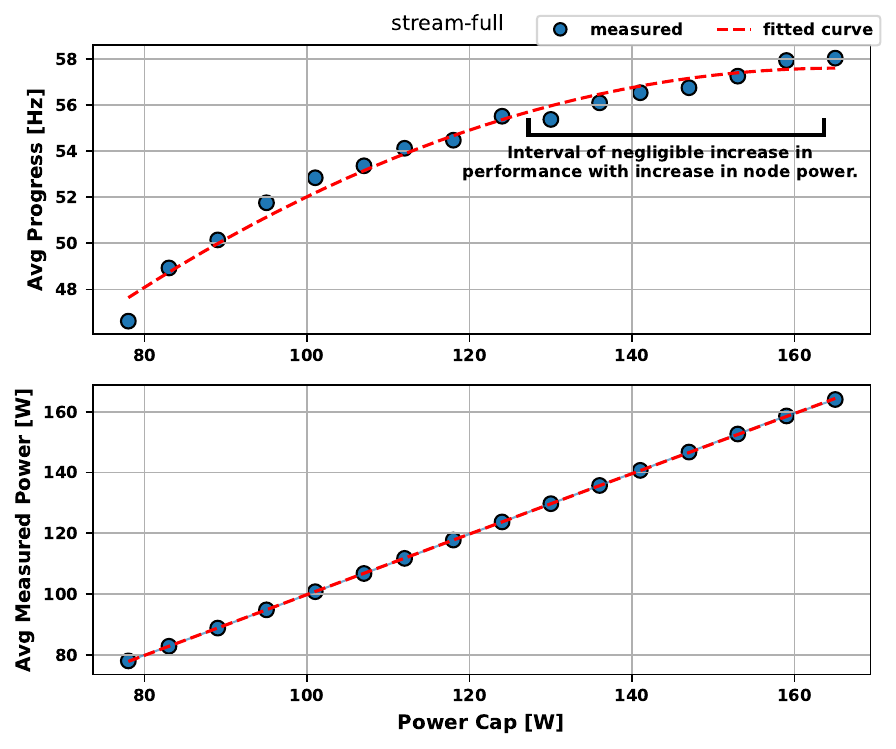}};

    \end{tikzpicture}
    \vspace{-0.7cm}
    \caption{Power vs instantaneous performance (progress) for the \texttt{stream-full} benchmark on an Intel Cascadelake node. Progress, defined in Eq.~\eqref{eq:progress-calculation}, is a proxy for the instantaneous performance of an application toward its scientific objective.}
    \label{fig:motivation_fig}
\end{figure}

Consider as an exemplar Figure~\ref{fig:motivation_fig}, which shows the instantaneous performance (instrumented measurement) of a memory-bound application with varying processor power caps and its corresponding power consumption.

\vspace{-0.3cm}
A closer scrutiny of the application's runtime performance and its corresponding CPU power consumption under different power caps reveals that power consumption in relation to these power caps does not linearly correlate with performance. Rather, there exist intervals on the performance curve where comparable performance levels are attainable at reduced power settings. Consequently, these intervals represent opportunities for optimization and are of particular interest to power and performance optimization controllers.

Modern processors with energy optimization technologies such as dynamic voltage and frequency scaling (DVFS)~\cite{4658633}, Intel's Running Average Power Limit (RAPL)~\cite{david2010rapl} and AMD's thermal design power cap~\cite{10.1145/3321551} have enabled application-oriented power optimization. 
Researchers have developed advanced power control strategies using these technologies, ranging from classical control theory to approaches based on machine learning (ML)~\cite{czarnul2019energy}. These methods can regulate power at a constant value, chosen according to hardware and application characteristics. Methods such as $DEPO$~\cite{KRZYWANIAK2023396} can even dynamically adjust power by observing changes in application performance. Despite these efforts, such approaches are either hardware-centric or application-specific: they require parameter changes whenever the application or hardware changes. 

In contrast, this paper presents the design of an application- and hardware-agnostic controller for energy-efficient computing using reinforcement learning (RL). By application agnostic, we mean that the controller does not rely on the specific nature or characteristics of the application. We  assume only that the applications are iterative in nature, such that an instrumented code segment can be placed within the most frequent iteration to measure the progress of the application toward its scientific objective. Similarly, hardware agnostic implies that the controller does not require any prior knowledge of the underlying system architecture or CPU characteristics for decision-making. 

RL, an online machine learning method, is generally used for control applications and, like any other ML algorithm, is computationally expensive to train. Therefore, researchers use surrogate models or digital twins of their environments~\cite{10305814} to train RL-agents when interactive learning is not possible. In such cases, the accuracy of the control objective that can be attained depends on the accuracy of the surrogate models used for learning. In high-performance computing (HPC) clusters and nodes, where application and hardware power-to-performance relationship changes across application and hardware architecture, generating a black-box model~\cite{WITT201933} using available datasets is difficult. Therefore, we introduce the use of an off-the-node training method for an RL agent using non-interactive or offline learning~\cite{levine2020offline}, with the data generated from existing arbitrary policies.
The training is offline in nature; in other words, the HPC node is never queried for evaluating an action during the training phase. This means offline RL requires a pre-collected dataset of state transitions, actions, and rewards, using which the agent can be trained elsewhere. By training the RL model offline, we reduce the computational burden on the HPC node and eliminate the need for surrogate models to train the agent.

We evaluate our approach by applying it to regulate the node-level power cap (PCAP) across multiple HPC nodes using 12 benchmark applications. The proposed algorithm is compared against several state-of-the-art power management systems~\cite{8903473,10.1007/978-3-319-58667-0_21,cerf2021sustaining}. Experimental results show that our scheme outperforms existing methods by reducing average energy consumption by 20\% with an average performance loss of 7.4\% and a worst-case performance degradation of 14\%. When compared with the manufacturer-provided \textit{on-demand} frequency governor---which dynamically adjusts core frequencies---our method demonstrates superior performance in 
reducing energy consumption while maintaining performance.

The key contributions of this paper are the following:
\begin{itemize}
\item We design an application and hardware agnostic power-capping controller for HPC systems that integrates lightweight online signals - including application heartbeats and hardware performance counters - to preserve performance while regulating energy use at runtime.
\item We demonstrate that offline reinforcement learning can be effectively used to learn such energy-efficiency control policies directly from previously collected datasets, enabling policy development without access to the live systems.
\item We validate our control policy on a range of well-established benchmarks characterized by varying arithmetic intensities and behaviors (phases).
\end{itemize}

Note that only for simplification purposes, we restrict the control design to multisocket CPU-only systems. Our methodology is applicable to other resources, and GPUs in particular will be the focus of future extensions of this work. Further, we posit that a single-node controller can then be integrated into a full-system solution. For example, an overprovisioned system could use systemwide power management to allocate available power to new resources, while using a controller such as ours to regulate the power consumption of allocated nodes.

The remainder of this paper is structured as follows: In Section ~\ref{sec:Related} we review existing algorithms in power and performance optimization. Section~\ref{sec:Background} presents the necessary background for this paper, including the RL algorithm, hardware and software stacks, and packages used in the experiments. 
Section~\ref{sec:Problem} formulates the problem. 
Section ~\ref{sec:Methodology} outlines the proposed algorithm and workflow in the context of compute nodes. In Section~\ref{sec:Setup} we present the experimental setup used in both training and evaluations. In Section ~\ref{sec:Results} we present the experimental design and implementation, along with the results and analysis.  Section ~\ref{sec:Conclusion} provides  a brief summary, the impact of this work, and future opportunities. 
\section{Related Work}\label{sec:Related}

Energy consumption of an HPC node running scientific applications has been the focus of substantial research in different domains of science and technology, including semiconductor design, architecture design, and application design. In this section we discuss relevant practices and techniques in the literature to reduce the energy consumption of HPC nodes given the hardware architecture and application. 

Power profiling is the conventional and convenient practice in the literature used to understand the application  phases ahead of its run. Ryan et al.~\cite{8323587} surveyed different methods used in the analysis of an application's runtime profile, where input power consumption changes. The authors classified the power-profiling methods into in-band and out-of-band, based on different approaches to measure the power consumption of a node. For instance, the jobwide energy consumption returned during or at the end of an application runtime, measured using external meters such as WattsUp! \cite{hirst2013watts} and 
WattProf~\cite{rashti2015wattprof}, is considered out-of-band, while the same data collected by analyzing the measurements from device-level integrated measurement sensors is considered in-band. From the perspective of an application during its execution, in-band measurements are more accurate and useful for dynamic power control. A survey conducted by Labasan et al.~\cite{labasan2016energy} explored state-of-the-art monitoring and controlling tools, such as Intel's RAPL, AMD's Application Power Management, IBM's EnergyScale, and the NVIDIA Management Library.

Simmendinger et al.~\cite{10740868} highlighted the importance of dynamic power
capping. Indeed, adjusting the power available dynamically can allow a management policy to deal with changes in applications' behaviors, as well as to redistribute
available power to other nodes or resources within a datacenter. 

Several works have explored system software capable of node-level energy
optimization. Most notably, the Energy Aware Runtime (EAR)~\cite{corbalan2019ear} is a software solution deployed in production systems capable
of regulating power at runtime using a model that detects MPI calls (phases) and
adjusts power caps accordingly. The Dynamic Energy-Performance Optimizer \cite{KRZYWANIAK2023396} takes care of GPU power control by using profiling tools to suggest power across the node. The dynamic power capping algorithm detects phase changes that were observed during
profiling runs, while it continues with static $PCAP$s for unobserved behaviors.
Similarly, the Bull Dynamic Power Optimizer \cite{8514915} uses instructions per
cycle (IPC) to dynamically adjust the core frequencies.

Our approach overcomes the application hardware-specific nature of these controllers, by being more amenable to control of unknown applications, while gathering application-agnostic performance data (heartbeats) for increased adaptability.

In another study, Cerf et al.~\cite{cerf2021sustaining} used a control theoretical approach  to identify optimal operating regions for compute nodes where performance decay is minimized under reduced power supply. They designed a classical proportional-integral-derivative controller to impose power caps  on RAPL actuators using performance feedback from the compute nodes. The controller was shown to track a user-defined performance set point.

The use of ML techniques that can simultaneously optimize performance and power is on the rise. Such methods have demonstrated increased efficacy in managing online job scheduling and hardware configurations \cite{chen2015distributed, 10571824}, offering broad applicability across various applications, where modeling the relationship between HPC configurations and performance is impractical.


The use of reinforcement learning is also not new in the control of HPC node energy efficiency. Wang et al.~\cite{wang2021online} proposed an online power control algorithm by observing the total instructions per cycle ($IPC$)  while running an application and thereby regulating core and uncore frequencies with online interactions during the execution of applications, to enhance learning. Ramesh et al. \cite{ramesh2019understanding} have already shown that $IPC$ is an unsuitable performance metric for HPC applications, as it fails to distinguish scientific progress from idle-time activity, both of which contribute to the total instruction count. The authors in~\cite{10305814} used RL trained on a mathematical model of a modified STREAM benchmark to regulate power of HPC nodes through RAPL actuators. But the process of generating a mathematical model for each  application's behavior to be trained is cumbersome. 


In this paper we formally introduce a dynamic power management algorithm that is independent of phase detectors and surrogate models and is capable of managing power across diverse hardware and applications to ensure optimal energy utilization at runtime.


\vspace{-0.2cm}
\section{Background}\label{sec:Background}
This section offers an overview of the concepts and technologies employed in this paper.
\subsection{Offline Reinforcement Learning}

Unlike supervised machine learning algorithms, RL algorithms do not depend on labeled data for training. Instead, an RL agent interacts with the system, evaluates the policy by testing the actions directly on the system, and trains by using the reward returned by the environment. 
RL is represented by using a Markov decision process, $\mathcal{M} := (\mathcal{S},\mathcal{A}, \mathcal{R}, \mathbf{T}, \gamma)$, where $\mathcal{S}$ is the set of states that are observable information of the environment; $\mathcal{A}$ is the action space; $\mathcal{R}$ is the reward function, which is a mapping from $(\mathcal{S} \times \mathcal{A} \times \mathcal{S}) \rightarrow \mathbb{R}$; $\mathbf{T}$ is the transition probability function, $\mathcal{S} \times \mathcal{A} \rightarrow \mathcal{S}$, following which the transition from the present state $s(t)$ to the next state $s(t+1)$ occurs under the action $a(t)$ while obtaining the reward $r(s,a)$; and $\gamma \in [0,1]$ is the discount factor, which determines how much importance the agent has to give to the long-term rewards rather than the immediate rewards.   The goal of any RL agent is to find a policy $\pi$, which is a mapping from each state in $\mathcal{S}$ to $\mathcal{A}$, that maximizes the total sum of discounted rewards (called the return or the $Q$ value) along the trajectory of experience. 

This online RL approach can be unsuitable, however, for certain real-world applications such as autonomous driving or robot control, where random actions cannot be applied to a live system because of safety or cost concerns. Parallel systems exhibit similar constraints: training an agent on a live system could cause disruption to users; it does not easily provide a reproducible and deterministic action-state response; and the training itself, because of its compute cost, could introduce further noise in the data. Therefore, in this paper we use offline RL~\cite{levine2020offline} to learn a policy from a static, precollected dataset. This dataset, which is considerably smaller than the full action-state domain under consideration, is gathered by using arbitrary policies. The offline training
algorithm then uses this dataset to learn an optimal policy without further system interactions. We note that no mainstream RL algorithm, including Conservative Q-learning (CQL), is capable of taking actions it has never seen during training. Therefore, we ensure that the dataset exhaustively samples the action space at least once across different applications and hardware combinations.


In an offline RL setting such as the Q-learning we use in this paper, we cannot evaluate an unaccounted state-action pair during the training phase. Instead, we rely on the ability of the Q-network to learn from the available dataset a value function, given any state. A common challenge associated with using such an offline dataset for training an RL agent is distributional shift. This refers to a situation where the distribution of states and actions being encountered during training differs from the real-world behavior, thereby adversely impacting the trained agent performance.

Here we use CQL \cite{levine2020offline} for training the RL-agent, a version of offline RL that regularizes the policy to avoid overestimation of the Q-values.By carefully choosing the learning parameters, we train an offline CQL agent from a previously collected dataset without overfitting. We recall the mathematical form of the loss function below for the reader’s understanding.

\begin{equation}
\begin{aligned}
    \min_Q \ & \overbrace{\max_{\mu} \Big(\mathbb{E}_{s\sim \mathcal{D}}\mathbb{E}_{a_t \sim \mu(a_t \vert s_t)}[Q(s_t,a_t)] - \mathbb{E}_{s_t,a_t \sim \mathcal{D}}[Q(s_t,a_t)]\Big)}^{\text{Conservative term}} \\
    & + \frac{1}{2\alpha}\mathbb{E}_{s_t,a_t,s_{t+1} \sim \mathcal{D}}[(Q(s_t,a_t) - r(s_t,a_t) \\& - \gamma \mathbb{E}_{a_{t+1} \sim \pi_{\theta}(a_{t+1} \vert s_{t+1})}[Q(s_{t+1},a_{t+1})])^2],
\end{aligned}
\label{eq:CQL}
\end{equation}
where $s_t,a_t$ represent, respectively, the present state and  action, $\mathbb{E}$ is the notation for expected value, $\mu$ is the current policy from which the action $a_t$ is chosen (initialized as random policy), $\mathcal{D}$ is the state distribution under policy $\mu$, $\alpha$ is the conservative learning factor that determines how the RL agent reacts to out of distribution data, and $\pi_{\theta}$ represents the parameterized target policy, where $\theta$ represents the weights of the Q-network. Referring to Equation~\ref{eq:CQL}, one can see that it consists of two parts. The first part minimizes overfitting or distributional shift by introducing a regularization term that constrains large Q-values while keeping the learning process unaffected. The second part corresponds to the conventional Q-learning objective, which aims to minimize the Bellman error.

\subsection{Measuring Application Performance through Hardware Performance Counters} \label{subsec:Applications_vs_perf}

Applications vary significantly with respect to their performance on a given hardware architecture.
The performance is influenced by hardware characteristics such
as memory bandwidth, CPU cores, and cache size. To understand this complex
application-hardware interaction, we capture a set of hardware performance
counters while running applications on the target system. We rely on the
Performance Application Programming Interface (PAPI)~\cite{mucci1999papi} to provide portable access to hardware performance counters.  PAPI offers easy access to a variety of counters through direct queries on the hardware model-specific registers. These measurements are representative of hardware-specific and application-specific data movement, or a combination of both.

To optimize energy efficiency while preserving performance, we focus on a small set of key performance metrics: total instructions issued ($TOT\_INS$), total clock cycles ($TOT\_CYC$), total last-level cache accesses ($L3\_TCA$), last-level cache misses ($L3\_TCM$), and cycles stalled because of resource contention ($RES\_STL$). We chose these key metrics for two reasons. First, these readings are available instantaneously and simultaneously during every sampling interval. Second, they are direct and sufficient indicators of the rate of data flow within and between the CPU cores and memory. Furthermore, because direct usage of these metrics would call for a large trained network and necessitate additional data processing, we follow insights from other research~\cite{wittmann2016chip,rane2014enhancing} in computing a set of \textit{derived metrics}---instructions per cycle ($IPC$), the ratio of stalled cycles to total cycles (often equated with CPU utilization, noted as $STL$), and last-level cache miss rate ($CMR$)---which facilitate a more detailed analysis of runtime behavior.  

\textbf{Key Remark:} We note, based on insights from previous works, that these metrics should not be viewed in isolation as definitive indicators of performance, since varying IPC values, for example, can still lead to the same execution time. Nonetheless, these derived metrics serve as valuable proxies to discern application characteristics, including its arithmetic intensity or whether the application is memory-bounded and thus can operate efficiently within a lower power budget.

\subsection{Application-Aware Performance Measurement} \label{subsec:Performance_Measurement}

We augment our monitoring of application behavior by adding a notion of \textit{progress}, or \textit{heartbeats}, to our system. This measure is obtained by a lightweight instrumentation of the applications under study.  Our approach follows the work of Ramesh et al.~\cite{ramesh2019understanding} and follows the instrumentation steps outlined by Cerf et al.~\cite{cerf2021sustaining}.
The role of this progress measurement is to highlight the potential for an application to be busy-waiting for some resource in the system and thus be active but not making useful use of compute power.

We redefine the progress measurement equation introduced in~\cite{ramesh2019understanding} by considering the number of heartbeats received between two sampling (control) instants. The progress at time instant $t_i$ is given by Equation~\eqref{eq:progress-calculation}:
\begin{equation}
\begin{aligned}
progress(t_i) = \underset{\forall k,t_k \in [t_{i-1},t_i]}{\text{median}}\bigg(\frac{N}{t_k-t_{k-1}} \bigg),
\end{aligned}
\label{eq:progress-calculation}
\end{equation}
where $N$ represents the number of heartbeats received between the time instants
$t_k$ and $t_{k-1}$. This measure of
progress can be understood as a proxy for monitoring the figure of merit
of an application at runtime. Equation \eqref{eq:progress-calculation} enables us to accurately measure progress while considering the time interval and the number of heartbeats received. It also allows us to limit the heartbeat-rate of instrumentation to avoid overflowing our infrastructure with messages, since  it is sufficient to accumulate a 
heartbeat counter on the application side and to only periodically send this
counter to our data collection and control infrastructure. 

\section{Problem Statement} \label{sec:Problem}
Given an application and a hardware node of an HPC cluster, the goal is to reduce energy consumption during the application's execution while allowing little to no degradation in performance. We formulate this problem as a multiobjective optimization problem (balancing energy and performance) that we linearize by instead minimizing the energy delay squared product ($ED^2P$).
$ED^2P$ is widely used among HPC researchers to measure the degradation in performance relative to the energy saved.
Compared with the energy-delay product (EDP) where equal importance is given to the optimization of energy and execution time, $ED^2P$ puts extra emphasis on maximizing the performance against minimization of energy. The formula is $ED^2P = E \times ET^2 $, where $ET$ represents execution time and $E$ represents energy. Both $ET$ and $E$ are  available only at the end of the application execution.

Referring to Figure~\ref{fig:motivation_fig}, we define the term performance per watt ($PPW$) as a measure of useful work done per unit power. For example, the $PPW$ corresponding to any operating point shown in Figure~\ref{fig:motivation_fig} is the square of the progress value per the measured power. This is a useful metric that is available d uring the application's execution. We pose the minimization of $ED^2P$ as a maximization of $PPW$ and $progress$. That is,
\begin{equation*}
\min \mathbb{E}[E \times ET^2] = \min \mathbb{E}[P_{avg}ET^3]
 = \min \mathbb{E}\Bigg[\int \frac{P_{avg,t_i}}{progress^2(t_i)}dt\Bigg],
 \end{equation*}
where $P_{avg}$ is the average power measured during the execution time and $P_{avg}(t_i)$ denotes the power measured during the $i^{th}$ time interval. In other words the goal of the above objective is to minimize the instantaneous values for $\frac{P_{avg,t_i}}{progress^2(t_i)}$ or otherwise maximize $\frac{progress^2(t_i)}{P_{avg,t_i}}$ = $PPW(t_i){progress(t_i)}$
\section{Methodology}\label{sec:Methodology}

\begin{figure}[htb]
    \centering
    \includegraphics[width=.9\linewidth,trim=5.5cm 6cm 3cm 1cm]{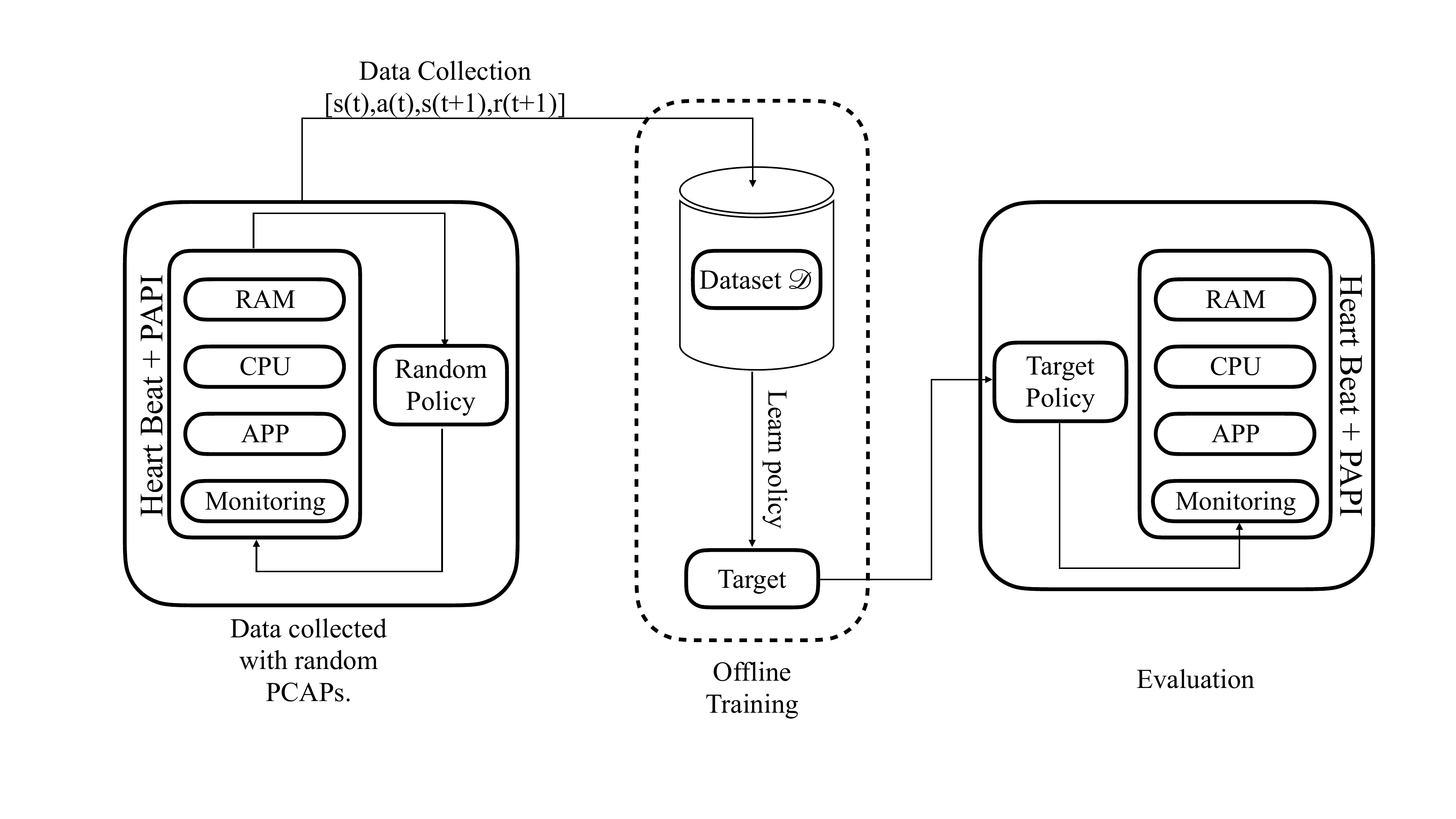}
    \caption{Flow diagram for offline reinforcement learning: data is collected using an arbitrary policy to train the agent without system access, and the trained agent is later deployed for evaluation. Note that the lines in the diagram are not closed to form a loop, indicating offline training and online testing.}
    \label{fig:flow-diagram}
\end{figure}

This section describes the methods, beginning with data collection, followed by the training of the RL agent and the deployment of the algorithm. To the best of our knowledge, this paper is the first to use an offline RL strategy to dynamically regulate the power of an HPC node. The different phases of the approach are depicted pictorially in Figure~\ref{fig:flow-diagram}. As shown in the figure, there are no closed loops between the RL agent and the evaluation process, meaning the agent is trained without feedback from direct interaction with the target system.

The initial phase involves data collection. The dataset consists of quadruples of state, action, next state, and reward, corresponding to different benchmarks (training/testing). Table \ref{tab:performance_metrics} presents the characteristics of various benchmarks used throughout this paper, including problem size, iterations, average $IPC$, average $CMR$, average $STL$, average progress, and average execution time ($ET$) for each kernel, calculated over 10 independent runs.
Unlike existing works \cite{10305814} that use a mathematical model to describe the relationship between application performance and power for training an RL agent, our approach learns directly from the dataset, which consists of performance and power data. By analyzing a set of precollected data, the RL agent aims to improve decision-making processes and optimize performance over the course of training iterations. 

It is crucial here to define the observables that constitute the states.
In our case, we represent the state of the RL agent using application behavior, identified through hardware performance counters. Similarly, the instantaneous performance of the application, measured through application instrumentation, and the ongoing power consumption are also integral to the state. These characteristics vary across applications and can be visualized by using the values provided in Table~\ref{tab:performance_metrics}.
Therefore, 
\begin{equation*}
s(t) = [progress(t), power(t), IPC(t), STL(t), CMR(t)].
\end{equation*}
The 
action is the CPU power supplied using the RAPL actuators, that is, $a(t) = PCAP(t)$. Upon observing $s(t)$ the system transitions to $s(t+1)$ under the action $a(t)$ following the transition dynamics under the policy.

Our objective, as defined in Section~\ref{sec:Problem}, is to identify the maximum efficiency point ($\max{PPW}$) of operation for the application on an HPC node, which will be represented by a suitable reward function. The reward function in RL helps the agent learn a policy that aligns with the optimization goal. In our case, the reward should be a function of application progress and power consumption, which together constitute $PPW$. The reward; function $r(t)$ is thus defined as
\begin{equation}
\begin{aligned}
reward(t+1) = \frac{progress^3(t+1)}{power(t+1)+1e^{-3}},
\end{aligned}
\label{eq:reward_function}
\end{equation}
where the reward is an explicit function of $progress(t+1)$ and $power(t+1)$ representing the progress and measured power, respectively. A small constant (0.001) is added to the denominator to avoid division by 0. The rest of the notations follow the earlier definitions. The overall structure of the reward function is designed to maximize the reward for improved progress while penalizing high actual power consumption. The $progress(t+1)$ term cubed in the numerator shows that more progress leads to a higher reward. The squared $PCAP(t)$ in the denominator suggests that the reward decrease is more than proportional to the increase in input power; thus, using significantly more input power than necessary reduces the reward substantially. 
We note that the progress values across different applications lie on varying scales. Therefore, we perform a normalization step prior to training to address this variability in the absolute reward values computed by using Equation ~\ref{eq:reward_function}. Specifically, we record the maximum and minimum reward values obtained for each application and rescale them uniformly to the range [-5,5]. This ensures that the rewards are well separated and not concentrated within a narrow interval. The chosen range can be adjusted depending on the dataset characteristics and the distribution of the reward function.
We note that in CQL the reward is part of the dataset and can be calculated before training, since it is a function of the next state and measured power.

The state variables include characteristics that provide insights into potential bottlenecks occurring during the application runtime. The reward function determines the efficiency or quality of $PCAP$ deployed to achieve the measured state. Therefore, training the RL agent allows it to learn a high-reward policy by observing the state $s(t)$ and resulting state $s(t+1)$. 

As pointed out in Section~\ref{sec:Background}, the CQL agent cannot extrapolate the action space. Therefore it is important to populate the dataset with actions sampled from the allowed action space. An ideal way to do so is by generating random action samples spanning the allowed $PCAP$ range. 

We use CQL for training the offline RL agent, which has the advantage of optimizing over the distributional shift, as detailed in Section~\ref{sec:Background}. The RL agent is defined by using a single Q-network, where the input size is equivalent to the state dimension and the output size corresponds to the cardinality of the action space. During training, the agent samples random batches of data from the dataset to train the agent. The Q-network weights are then updated to learn the action policy that returns the maximum reward from that batch of data. These iterations continue until convergence of the loss function given in Equation~\ref{eq:CQL}. 


We evaluate the trained model by running it alongside test applications that the training phase has not encountered before, as well as some applications with phases (to evaluate adaptability). During evaluation, the Q-network computes the action $Q$-values corresponding to the observed state provided by the daemon process and takes the greedy action, that is, $PCAP = \arg\max_a(Q(s, a))$. This $PCAP$ is deployed by using the $RAPL$ actuators and is maintained at a steady throughput throughout the sampling interval. The sampling interval plays an important role in determining the adaptability of the controller during actuation. In our experiments we have intentionally kept the sampling interval smaller to detect minute changes in performance.

\section{Experimental Setup}\label{sec:Setup}

All the experiments in this paper, including data collection and evaluation, were performed on a bare-metal Cascadelake node from Chameleon Cloud~\cite{keahey2019chameleon}, equipped with Intel Xeon Gold 6240R processors. The node has 2 processors, each with 24 cores and 2 hyperthreads per core, and 192 GiB of memory. The L3 cache size is 35.75 MiB per processor.

Provisioning of these nodes included a fully installed and configured Global Extensible Open Power Manager (GEOPM)~\cite{eastep2017global} service (version $3.1.0$) to enable reading and writing of power sensors, but without GEOPM being configured to provide any power control policy itself. PAPI version $6.0.0$ was used to access hardware performance counters. A custom software infrastructure handles the collection of data from GEOPM, PAPI, and application heartbeats, while also providing the basis for our control policy deployment.

\subsection{Benchmarks}

We use a collection of well-established benchmarks to represent the behaviors of various applications and their response to power capping. The benchmarks
listed here are used both for training and for validating our control agent, with some of them being kept out of our training dataset and used  only for validation. By using these well-established benchmarks, we hope to train our model to recognize and predict optimal power management strategies that will be applicable to a wide range of application behaviors.


\begin{table*}[htb]
    \centering
    \caption{Detailed description of our benchmarks, including their type, parameters, and average performance metrics (uncapped).}
    \label{tab:performance_metrics}
    \resizebox{\linewidth}{!}{%
        \begin{tabular}{|c|c|c|c|c|c|c|c|c|c|}
            \hline
            \textbf{Benchmark} & \textbf{Prob Size} & \textbf{Iter.} & \textbf{Avg. IPC} & \textbf{Avg. CMR} & \textbf{Avg. STL} & \textbf{Avg. Prog.} \textbf{[\unit{Hz}]} & \textbf{Avg. ET}\textbf{[\unit{s}]} & \textbf{Avg. Energy} \textbf{[\unit{kJ}]} & \textbf{Ari. Int. (FLOPs/Byte)} \\
            \hline
            {\textbf{STREAM SCALE}}  & 33554432 & 10000 & 0.20 & 0.89 & 0.84 & 285.20 & 34.40 & 5.59 & 1.62 \\
            \hline
            {\textbf{STREAM TRIAD}}  & 33554432 & 10000 & 0.18 & 0.94 & 0.83 & 200.03 & 50.80 & 7.93 & 1.41 \\
            \hline
            {\textbf{NPB-EP}}  & Class-W & 1000 & 0.57 & 0.13 & 0.48 & 13.61 & 73.80 & 10.33 & 53.99 \\
            \hline
            {\textbf{NPB-IS}}  & Class-B & 1000 & 0.50 & 0.86 & 0.68 & 8.44 & 117.33 & 18.99 & 1.21 \\
            \hline
            {\textbf{NPB-MG}}  & Class-B & 1000 & 0.47 & 0.809 & 0.54 & 40.3 & 22.23 & 55.5 & 1.45 \\
            \hline
            {\textbf{NPB-FT}}  & Class-B & 500 & 0.82 & 0.43 & 0.296 & 13.23 & 35.10 & 45.47 & 1.53 \\
            \hline
            \hline
            {\textbf{STREAM-ADD}}  & 33554432 & 10000 & 0.15 & 0.94 & 0.85 & 201.06 & 50.40 & 7.84 & 1.36 \\
            \hline
            {\textbf{STREAM-COPY}}    & 33554432 & 10000 & 0.17 & 0.89 & 0.84 & 282.43 & 35.50 & 5.66 & 1.63 \\
            \hline
            {\textbf{STREAM-FULL}}  & 33554432 & 10000 & 0.16 & 0.93 & 0.70 & 49.31 & 222.40 & 34.99 & 1.45 \\
            \hline
            {\textbf{STREAM-PHASE}}  & 33554432 & 10000 & 0.17 & 0.91 & 0.88 & 240.16 & 83.00 & 13.48 & 1.45 \\ 
            \hline
            {\textbf{NPB-CG}}  & Class-C & 1000 & 0.49 & 0.37 & 0.62 & 9.97 & 114.00 & 14.74 & 3.61 \\
            \hline
            {\textbf{NPB-BT}}  & Class-C & 1000 & 0.97 & 0.87 & 0.30 & 6.22 & 166.00 & 23.168 & 0.54 \\
            \hline
        \end{tabular}
    }
\end{table*}

\textbf{STREAM}~\cite{McCalpin1995} is a well-known benchmark designed to
evaluate the memory bandwidth of a system. We consider six versions of this
benchmark, corresponding to its four inner kernels (add, copy, scale, and triad)
with varying arithmetic intensity, as well as one version (\texttt{full}) that
executes all four kernels as part of its outer loop. The last version, with phases,
repeats the full benchmark a number of times, resulting in changes in progress
rate and behavior across its execution. For all these versions, the input
problem size (the size of the working arrays in memory) has a direct impact on
the performance and is chosen here so that these arrays do not fit in the last
level cache on the system under study.
    
\textbf{NAS} Parallel Benchmarks (NPB) are a collection of programs
created to benchmark the performance of parallel processors using kernels that
are representative of scientific applications. 
We include six of the benchmarks here: Embarrassingly Parallel (EP) for being a
representative of a compute-bound application, Integer Sort (IS) for its focus
on random memory accesses, Fourier Transform (FT) for its straining of the cache for
memory access, Multi-Grid (MG) for its computational intensity, Conjugate
Gradient (CG) for memory-intensive and irregular communication patterns, and
Block Tridiagonal (BT) for its compute intensive and irregular communication.
Our implementation is based on the NPB-3.4-OpenMP version. 

We modified these benchmarks to be iterative so that application progress is more easily identifiable: When not already a part of the benchmark, an outer loop is introduced, repeating their kernels a configurable amount of time. Each outer loop also contains our instrumentation for heartbeats, resulting in each  benchmark reporting one heartbeat per iteration. Note that the progress rate of each benchmark is different and depends on the runtime of its kernel.

Since this paper is concerned with demonstrating energy efficiency, we select each benchmark problem size to be bigger than caches, with the goal of exhibiting performance trade-offs between CPU power capping (and the resulting frequency variation) and stable memory performance.
These problem size values are provided in Table~\ref{tab:performance_metrics}. The table also provides information on the arithmetic intensity for each application, defined as the ratio between total computational operations  and total instructions, which serves as a measure of computational diversity among the applications. A double line  in the table separates applications that are part of the training dataset (top) and those used only during validation (bottom).

\subsection{Data Collection and Offline Training}


We collect our training dataset by running these benchmarks under an arbitrary
control policy, sampling all the measurements that  our state vector comprises
during these runs. To ensure that the training dataset fits the conditions under
which the trained controller will operate, we set up the experiments  to sample
at the same rate as the control will activate: 1 Hz (once per second). In this
paper we use a random control policy during data collection: $PCAP$ values for RAPL,
chosen randomly from a discretization of the action space of allowed power caps
for the system. Indeed, while the system allows for a power capping range of 78
\unit{\watt} to 165 \unit{\watt}, we apply only 16 values distributed uniformly
in that range. By reducing the action state to such a small and finite set of
actions, we avoid the need for a much larger network (in the output dimension)
to train. This choice also significantly reduces the size of the  training dataset
while preserving the controller’s performance. We ensure in our data collection
that all 16 possible power caps are sampled, since CQL, like most RL algorithms,
does not extrapolate on the action space observed. Nevertheless, not all
transitions between two actions corresponding to all state observations are
present in the dataset. 

This collection strategy is repeated for each of the benchmarks selected as part
of the training set for the problem size and iterations given in  Table
\ref{tab:performance_metrics}. Rewards are computed from these state vectors and
appended to the dataset and normalized as explained in Section~
\ref{sec:Methodology}.


The resulting dataset consists of 1,291 data points ($s(t), a(t), r(t+1),
s(t+1)$),  on which offline training is performed. Since the action space is
finite, we use a Q-network consisting of output dimensions equivalent to the
cardinality of the action space (i.e., 16). The input dimension is the dimension
of the state space or observation space, which is 5. The intermediate layers
consist of 10 neurons each. A total of 10,000 iterations, with a discount factor
$\gamma = 0.9$ and the CQL constant $\alpha = 0.1$, was used for training, with
a buffer size of 128. It is recommended to keep the $\alpha$ at larger values
(>0.3) for larger datasets to compensate for out-of-distribution observations.
The number of iterations is chosen based on the time it takes for the Q-network
to converge.

\textit{All the software used is open-source and will be linked in the paper for the camera-ready version of this paper.}

\begin{figure*}[htb]
\centering
    \vspace{-0.1cm}
    \includegraphics[width=0.8\linewidth]{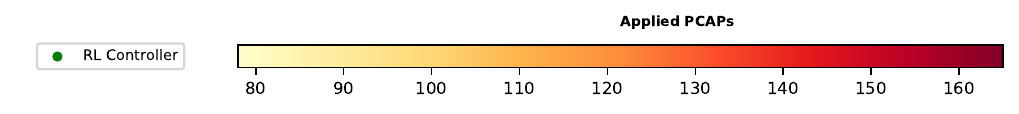} \\
    \vspace{-0.15cm}
    \includegraphics[width=0.95\linewidth, trim=0cm 0cm 0 0]{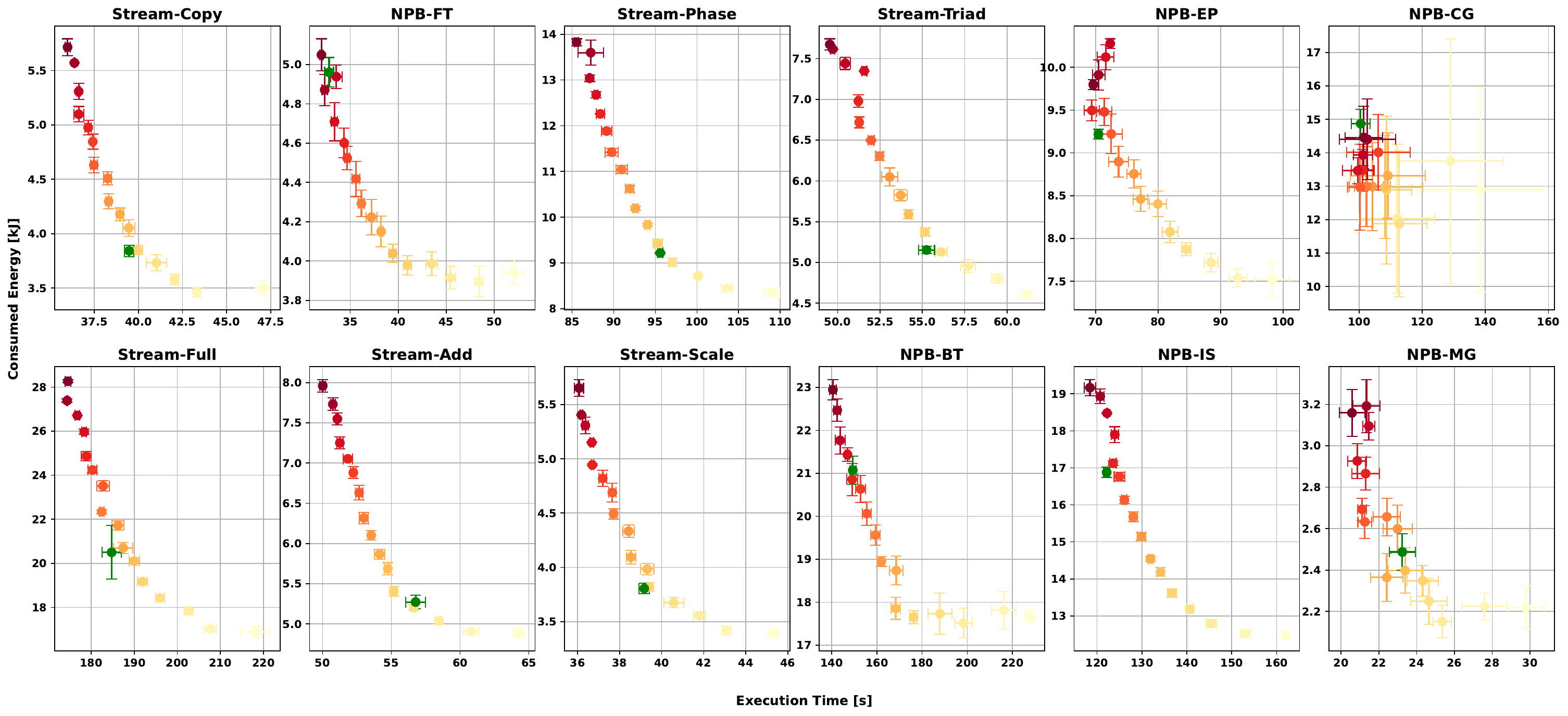}
\caption{Execution time vs. energy for twelve benchmarks. Six applications were unseen during training (Table \ref{tab:repeatability}). Yellow-red gradients show fixed $PCAP$ values; the green dot shows the proposed controller. Each point is averaged over five runs.}
\label{fig:comparison}
\end{figure*}
\section{Empirical Evaluations}\label{sec:Results}
We now evaluate the proposed controller by deploying it to regulate the node's power while executing both training and testing applications. Later, we also compare our method with state-of-the-art controllers.

\subsection{Evaluation of the Proposed Framework}
We assess the execution time and energy consumption in comparison with the performance of an uncapped system. In all the testing scenarios consisting of different applications, the agent evaluates states that it has never encountered during training. More out-of-distribution states are likely to appear when the proposed controller is used to regulate the power of testing applications.

For testing, we deploy our infrastructure, the trained agent, and the application together on the node of interest. The infrastructure is designed to collect the measurements ($progress$, $power$, and $PAPI$) and to provide them to the controller, which, upon postprocessing and observing the required states, makes an informed decision. The resulting action is then translated into RAPL control values and applied on the system through GEOPM.

We first test the controller on the applications used for training, namely, the STREAM TRIAD, STREAM SCALE, NPB-IS, and NPB-EP benchmarks, which are selected to provide a large range of memory boundedness behavior. We then collect the runtime metrics and calculate the execution time and total energy consumed during the runtime for each experiment. Furthermore, we assess the performance and energy consumption of the controller with applications it has never encountered during training. For this analysis, we use the testing applications listed in Table \ref{tab:performance_metrics}.
For comparison, we generate a Pareto curve of energy versus execution time for all the applications used in both training and testing, as shown in Figure~\ref{fig:comparison} using various static power caps. 

Figure~\ref{fig:comparison} illustrates the relationship between energy consumption and execution time across our experiments. The comparison shows that the RL controller (green dot) consistently regulates the node's operating region between the maximum and minimum execution extremes, particularly for memory-bound applications. For compute-bound applications, the green dot typically appears on the darker end of the color bar, indicating operation under higher $PCAP$ settings. Notably, the green dot always lies on or outside the Pareto curve, demonstrating that the RL controller achieves improved or comparable performance under static workloads. This plot also highlights that the RL controller does not introduce any noticeable computational overhead on the node.

Table~\ref{tab:repeatability} presents statistics related to the repeatability of our experiments. Together with Figure~\ref{fig:comparison}, it demonstrates that the proposed controller achieves significantly greater energy savings than any corresponding loss in performance. On average, the controller reduced energy consumption by 20.2\%, with only a 7.5\% decrease in performance compared with uncapped execution.

Some exceptions can be observed in the $NPB-CG$ and $NPB-BT$ cases, which do not seem to be controlled as efficiently or as stably as the other benchmarks. We suspect that their characteristics, including a smaller average progress than most other benchmarks, might represent a challenge when not including such applications as part of the training dataset. A more exhaustive search for the ideal training set, however,  is  quickly prohibitive as the number of applications increases. Nevertheless, these benchmarks do not result in a performance degradation higher than that of the other benchmarks.

\begin{table*}[htb]
\caption{Execution time and energy of all benchmarks under three power cap strategies: Min $PCAP$, Max $PCAP$, and the proposed RL agent. Results show mean and standard deviation over 5 runs, including percentage execution-time degradation and energy reduction of RL relative to Max $PCAP$.}
\begin{center}
\resizebox{\textwidth}{!}{%
\begin{tabular}{|c|c|c|c|c|c|c|c|c|c|c|c|c|c|c|}
\hline
\multirow{3}{*}{\textbf{Benchmark}} & \multicolumn{4}{c|}{\textbf{Min $PCAP$}} & \multicolumn{4}{c|}{\textbf{Max $PCAP$}} & \multicolumn{6}{c|}{\textbf{RL}} \\
\cline{2-15}
& \multicolumn{2}{c|}{\textbf{Execution Time}} & \multicolumn{2}{c|}{\textbf{Energy}} & \multicolumn{2}{c|}{\textbf{Execution Time}} & \multicolumn{2}{c|}{\textbf{Energy}} & \multicolumn{2}{c|}{\textbf{Execution Time}} & \multicolumn{2}{c|}{\textbf{Energy}} & \shortstack{\scriptsize{ET Degradation}  (\%)} & \shortstack{\scriptsize{Energy Saved}  (\%)} \\
\cline{2-15}
& $\mu$[\unit{s}] & $\sigma$ & $\mu$[\unit{kJ}] & $\sigma$ & $\mu$[\unit{s}] & $\sigma$ & $\mu$[\unit{kJ}] & $\sigma$ & $\mu$[\unit{s}] & $\sigma$ & $\mu$[\unit{kJ}] & $\sigma$ & (\%) & (\%) \\
\hline
\textbf{STREAM SCALE} & 37.73 & 2.61 & 4.17 & 0.61 & 34.33 & 0.47 & 5.65 & 0.08 & 37.33 & 0.47 & 3.81 & 0.05 & 8.74 & 32.61 \\
\hline
\textbf{STREAM TRIAD} & 52.86 & 3.26 & 5.83 & 0.88 & 47.67 & 0.47 & 7.67 & 0.07 & 53.67 & 0.47 & 5.15 & 0.05 & 12.59 & 32.90 \\
\hline
\textbf{NPB-EP} & 78.74 & 8.76 & 8.54 & 0.84 & 67.67 & 0.47 & 9.80 & 0.06 & 68.67 & 0.47 & 9.22 & 0.06 & 1.48 & 5.92 \\
\hline
\textbf{NPB-IS} & 134.85 & 11.89 & 14.67 & 1.90 & 116.67 & 1.25 & 19.17 & 0.22 & 120.67 & 0.94 & 16.88 & 0.14 & 3.43 & 11.94 \\
\hline
\textbf{NPB-MG} & 22.75 & 2.60 & 2.60 & 0.34 & 19.80 & 0.40 & 3.18 & 0.09 & 22.67 & 0.94 & 2.49 & 0.09 & 14.47 & 21.70 \\
\hline
\textbf{NPB-FT} & 37.52 & 5.81 & 4.30 & 0.35 & 30.60 & 0.49 & 5.05 & 0.08 & 31.33 & 0.47 & 4.96 & 0.08 & 2.39 & 1.78 \\
\hline\hline
\textbf{STREAM ADD} & 53.69 & 3.88 & 5.92 & 0.86 & 48.33 & 0.47 & 7.96 & 0.08 & 55.00 & 0.82 & 5.27 & 0.08 & 13.80 & 33.74 \\
\hline
\textbf{STREAM COPY} & 38.09 & 2.92 & 4.20 & 0.60 & 34.67 & 0.47 & 5.72 & 0.08 & 37.67 & 0.47 & 3.84 & 0.05 & 8.66 & 32.91 \\
\hline
\textbf{STREAM FULL} & 189.77 & 12.24 & 20.86 & 3.13 & 172.33 & 0.47 & 28.27 & 0.07 & 183.00 & 2.52 & 20.50 & 1.22 & 6.20 & 27.47 \\
\hline
\textbf{STREAM PHASES} & 93.31 & 6.26 & 10.22 & 1.52 & 84.33 & 0.47 & 13.83 & 0.07 & 94.00 & 0.00 & 9.22 & 0.01 & 11.46 & 33.34 \\
\hline
\textbf{NPB-CG} & 109.96 & 15.83 & 13.79 & 2.04 & 101.00 & 5.97 & 15.08 & 0.93 & 100.17 & 2.91 & 14.87 & 0.43 & -0.82 & 1.37 \\
\hline
\textbf{NPB-BT} & 169.00 & 24.89 & 19.43 & 1.68 & 140.00 & 1.67 & 22.99 & 0.25 & 149.00 & 1.41 & 21.07 & 0.33 & 6.43 & 8.34 \\
\hline
\end{tabular}
}
\label{tab:repeatability}
\end{center}
\end{table*}

\begin{figure*}[htb]
\centering
    \vspace{-0.1cm}
    \includegraphics[width=0.8\linewidth]{Figures/legend_only.pdf} \\
    \includegraphics[width=0.95\linewidth, trim=0cm 0cm 0 0]{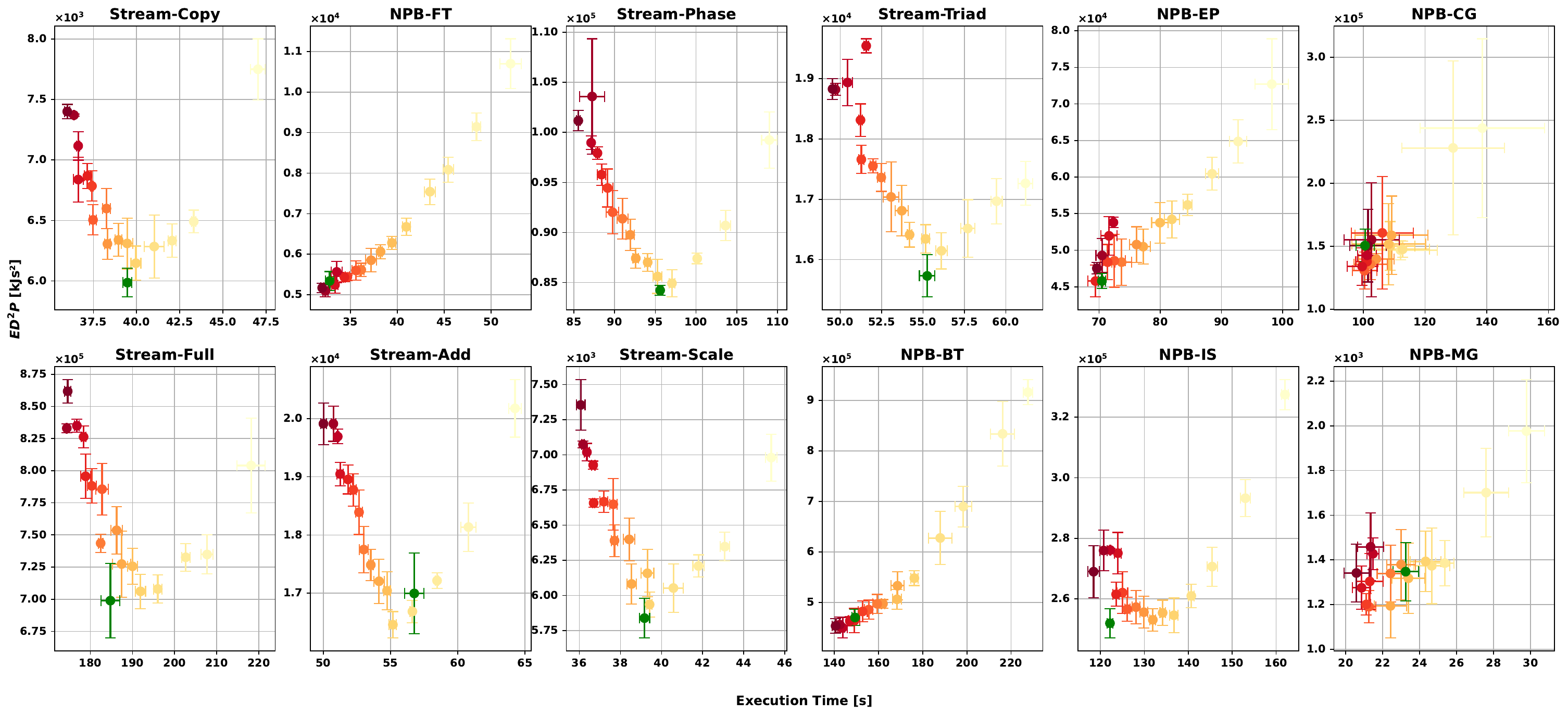}
\caption{Mean and standard deviation of $ED^2P$ vs. execution time for the experiments in Fig.~/ref{fig:comparison}. The proposed controller consistently maintains lower average $ED^2P$ across applications.}
\label{fig:ED2P}
\end{figure*}

To understand the impact of the proposed controller in optimizing the target objective, we plot the ED2P values corresponding to each of these experiments. The results appear in Figure~\ref{fig:ED2P}. 
The results show that the proposed controller was able to regulate power across different application scenarios, often where the ED2P value was close to its global minimum. Compared with the uncapped power setting, the controller gave better results in all cases. Since there is no known optimal closed-loop controller, it is useful to compare against the uncapped case.


\sisetup{
  scientific-notation=true,
  round-mode=figures,
  round-precision=3,
  detect-weight=true,
  detect-inline-weight=math
}
\newcommand{\bestg}[1]{\cellcolor{green!25}\textbf{#1}}
\newcommand{\worstr}[1]{\cellcolor{red!15}#1}

\begin{table*}[t]
\begin{center}
\resizebox{\textwidth}{!}{%
\begin{tabular}{|c|ccc|ccc|ccc|ccc|ccc|}
\hline
\multirow{3}{*}{\textbf{Benchmark}} &
\multicolumn{3}{c|}{\textbf{DVFS Controller}} &
\multicolumn{3}{c|}{\textbf{PI-Global Model}} &
\multicolumn{3}{c|}{\textbf{PI-App-Specific Model}} &
\multicolumn{3}{c|}{\textbf{Ondemand Governor}} &
\multicolumn{3}{c|}{\textbf{RL (Ours)}} \\
\cline{2-16}
& \shortstack{\scriptsize Energy\\Saved (\%)} &
  \shortstack{\scriptsize Perf\\Degrad. (\%)} & \textbf{ED$^{2}$P} &
  \shortstack{\scriptsize Energy\\Saved (\%)} &
  \shortstack{\scriptsize Perf\\Degrad. (\%)} & \textbf{ED$^{2}$P} &
  \shortstack{\scriptsize Energy\\Saved (\%)} &
  \shortstack{\scriptsize Perf\\Degrad. (\%)} & \textbf{ED$^{2}$P} &
  \shortstack{\scriptsize Energy\\Saved (\%)} &
  \shortstack{\scriptsize Perf\\Degrad. (\%)} & \textbf{ED$^{2}$P} &
  \shortstack{\scriptsize Energy\\Saved (\%)} &
  \shortstack{\scriptsize Perf\\Degrad. (\%)} & \textbf{ED$^{2}$P} \\
\cline{2-16}
&  &  &  &  &  &  &  &  &  &  &  &  &  &  &  \\
\hline

\textbf{NPB-BT} &
\worstr{-13.50} & \worstr{13.90} & \worstr{\num{6.63436e5}} &
\worstr{-98.43} & \bestg{-0.21} & \num{8.90421e5} &
-5.12 & 7.14 & \num{5.43782e5} &
-13.50 & 13.90 & \num{6.63436e5} &
\bestg{8.34} & 6.43 & \bestg{\num{4.67824e5}} \\
\hline

\textbf{NPB-CG} &
-7.74 & -1.44 & \num{1.60989e5} &
\worstr{-13.56} & \worstr{9.57} & \worstr{\num{2.09714e5}} &
2.23 & 0.99 & \num{1.53389e5} &
-11.47 & 11.17 & \num{2.11919e5} &
\bestg{1.37} & \bestg{-0.83} & \bestg{\num{1.49226e5}} \\
\hline

\textbf{NPB-EP} &
-0.80 & -1.56 & \num{4.38312e4} &
\worstr{-124.00} & \worstr{16.87} & \worstr{\num{1.37298e5}} &
6.33 & 4.43 & \num{4.58408e4} &
-9.97 & 15.63 & \num{6.59907e4} &
\bestg{5.93} & \bestg{1.47} & \bestg{\num{4.34663e4}} \\
\hline

\textbf{NPB-FT} &
-5.70 & \worstr{36.42} & \num{9.30167e3} &
\worstr{-123.59} & 16.31 & \num{1.43021e4} &
0.02 & 13.07 & \num{6.04468e3} &
\bestg{10.51} & \bestg{4.65} & \bestg{\num{4.63399e3}} &
1.78 & 2.40 & \num{4.86997e3} \\
\hline

\textbf{NPB-IS} &
-8.17 & \worstr{23.42} & \num{4.29962e5} &
\worstr{-127.69} & 17.41 & \worstr{\num{8.18978e5}} &
10.45 & 4.28 & \num{2.54109e5} &
-5.54 & 7.65 & \num{3.19145e5} &
\bestg{11.96} & \bestg{3.43} & \bestg{\num{2.45745e5}} \\
\hline

\textbf{NPB-MG} &
-11.04 & 27.19 & \num{2.23950e3} &
\worstr{-162.41} & \worstr{33.39} & \worstr{\num{5.82111e3}} &
6.23 & \bestg{-1.32} & \bestg{\num{1.13841e3}} &
-12.72 & 15.20 & \num{1.86513e3} &
\bestg{21.79} & 14.48 & \num{1.27779e3} \\
\hline

\textbf{STREAM FULL} &
7.76 & 1.50 & \num{7.97888e5} &
\worstr{-47.49} & \worstr{19.33} & \worstr{\num{1.76332e6}} &
9.21 & 4.06 & \num{8.25458e5} &
1.23 & 0.29 & \num{8.34002e5} &
\bestg{27.50} & \bestg{6.19} & \bestg{\num{6.86421e5}} \\
\hline

\textbf{STREAM TRIAD} &
7.66 & 12.86 & \num{2.05003e4} &
\worstr{-128.48} & \worstr{20.24} & \worstr{\num{5.75778e4}} &
11.17 & 10.48 & \num{1.88981e4} &
-4.44 & \bestg{6.45} & \num{2.06296e4} &
\bestg{32.82} & 12.58 & \bestg{\num{1.48407e4}} \\
\hline

\textbf{STREAM ADD} &
10.97 & 13.96 & \num{2.14976e4} &
\worstr{-147.56} & \worstr{27.35} & \worstr{\num{7.46519e4}} &
\bestg{35.17} & \bestg{4.60} & \bestg{\num{1.31873e4}} &
0.60 & 2.87 & \num{1.95567e4} &
33.77 & 13.80 & \num{1.59476e4} \\
\hline

\textbf{STREAM SCALE} &
9.61 & 15.77 & \num{8.06690e3} &
\worstr{-131.15} & \worstr{22.24} & \worstr{\num{2.29990e4}} &
12.17 & 8.75 & \num{6.91671e3} &
-0.84 & \bestg{6.49} & \num{7.61543e3} &
\bestg{32.65} & 8.75 & \bestg{\num{5.30353e3}} \\
\hline

\textbf{STREAM COPY} &
5.63 & 16.86 & \num{8.86046e3} &
\worstr{-143.02} & \worstr{28.76} & \worstr{\num{2.77001e4}} &
11.65 & 11.53 & \num{7.55612e3} &
1.23 & \bestg{4.82} & \num{7.46148e3} &
\bestg{32.83} & 8.64 & \bestg{\num{5.45115e3}} \\
\hline

\textbf{STREAM PHASES} &
11.15 & 14.51 & \num{1.14581e5} &
\worstr{-65.80} & \worstr{24.02} & \worstr{\num{2.50832e5}} &
2.20 & 7.12 & \num{1.10376e5} &
-0.35 & \bestg{3.38} & \num{1.05472e5} &
\bestg{33.35} & 11.47 & \bestg{\num{8.14447e4}} \\
\hline\hline

\textbf{Average} &
0.49 & 14.45 & \num{1.90096e5} &
\worstr{-109.43} & \worstr{19.61} & \worstr{\num{3.56135e5}} &
8.48 & 6.26 & \num{1.65558e5} &
-3.77 & 7.71 & \num{1.88477e5} &
\bestg{20.34} & \bestg{7.40} & \bestg{\num{1.43485e5}} \\
\hline
\end{tabular}
}
\caption{Controllers compared with uncapped (Max PCAP). Only rowwise extremes are colored: \textbf{green} = best (max Energy Saved, min Perf Degradation, min ED$^{2}$P); \textbf{red} = worst (min Energy Saved, max Perf Degradation, max ED$^{2}$P). Lower ED$^{2}$P is better.}
\label{tab:controllers_vs_uncapped_extremes}
\end{center}
\end{table*}
\subsection{Comparison with Existing Methods}
We compare our proposed method against several state-of-the-art approaches. The baseline scheme is defined as uncapped power consumption, with the CPU frequency governor set to \textit{performance}. The other methods compared against this baseline are described here.

\textbf{PI Controller}~\cite{cerf2021sustaining}: This implementation of PCAP control employs a classical proportional–integral (PI) controller with user-defined set points to specify the minimum allowable performance degradation. It uses both application- and hardware-based performance measurements to construct a model that serves as the basis for controller design. We use this controller in two variants: one based on a global model for all applications and hardware configurations and another that employs application- and hardware-specific models tailored to each case.

\textbf{\textit{Ondemand} CPU Governor}~\cite{wysocki2017cpufreq}: This scheme relies on the manufacturer-provided \textit{ondemand} governor to dynamically adjust CPU frequencies during execution. Here, applications run without explicit power capping, with power managed implicitly by the governor based on workload demand.

\textbf{RL-Based DVFS Control}~\cite{9465707}: This method uses an online reinforcement learning approach to regulate core frequency through dynamic voltage and frequency scaling (DVFS). The RL agent interacts with the system at runtime to train its model and determine suitable core and uncore power settings for the node executing the application. This method, however, relies solely on IPC as a measure of performance, which has been shown to be an unreliable indicator of scientific progress~\cite{ramesh2019understanding}.

For the PI controllers, we tuned application-specific parameters of the common mathematical model from ~\cite{cerf2021sustaining} based on the performance data collected for offline training. Then, each PI controller regulates node power under the same experimental conditions as the RL controller, for each application.
Because PI controller design depends heavily on the underlying application and hardware characteristics, a separate parametrization was required for each case. To evaluate generalization, we also developed a single, combined global PI shared across all applications, fitting the model as best as possible across applications. However, performance and power regulation using this global model is significantly worse, primarily because of model inaccuracy, which led to noisy actuation and unstable control behavior.

Similarly, we trained a DVFS-based RL controller using the same dataset employed for training our proposed offline RL method but considering only the instructions per cycle  as the performance metric. Unlike the original paper ~\cite{9465707} this training was performed offline since online training on the HPC node can interfere with application runtime and introduce noisy measurements. The resulting DVFS controller was evaluated under the same testing conditions as the proposed RL controller. For the \textit{ondemand} governor, the Linux kernel autonomously selected the optimal core frequency during execution while keeping the node power uncapped.

Among all the compared methods, the PI controller with a global model performed the worst, exhibiting frequent PCAP oscillations that resulted in higher energy consumption and longer execution times. The application-specific PI controller performed better, achieving the second-lowest average ED$^2$P among the baseline methods. The DVFS-based RL controller performed well on compute-bound applications, as its frequency regulation aimed to maximize IPC---an effective proxy for such workloads. It remained a close competitor to the proposed RL controller on memory-bound applications, outperforming it only on the STREAM FULL benchmark.

\begin{figure}[htb]
    \centering
    \begin{subfigure}[b]{0.48\textwidth}
        \centering
        \includegraphics[width=\textwidth]{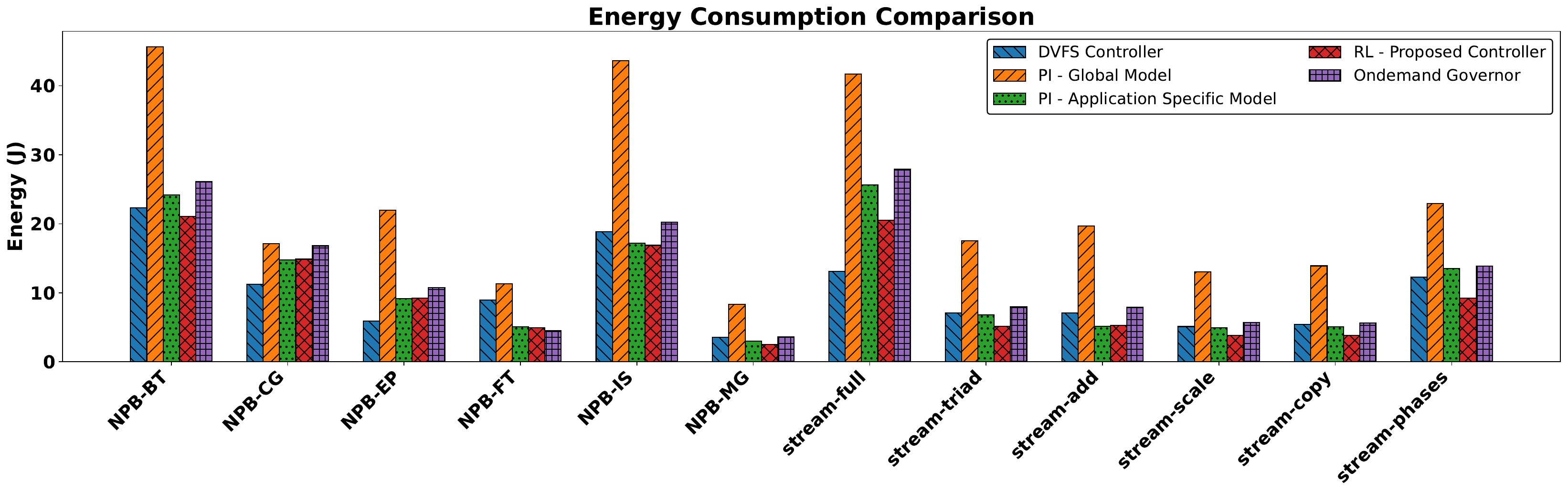}
        \caption*{}
        \label{fig:PI-controller}
    \end{subfigure}\hfill
    \vspace{-1cm}
    \begin{subfigure}[b]{0.48\textwidth}
        \centering   \includegraphics[width=\textwidth]{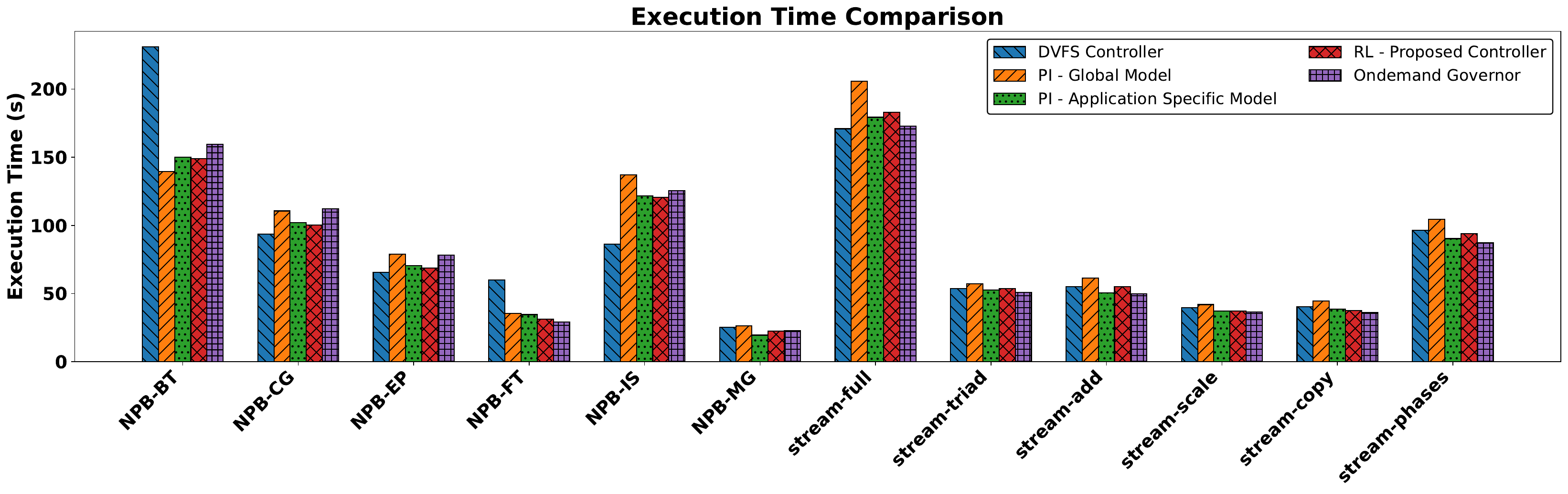}
        \caption*{}
        \label{fig:RL-controller}
    \end{subfigure}
    \vspace{-1.5cm}
    \caption{Aggregate comparison between the proposed RL-based power-capping controller and four baselines: global PI, application-specific PI, DVFS, and the \textit{ondemand} governor.
    }
    \label{fig:comparison_PI}
\end{figure}

As shown in Figure~\ref{fig:comparison_PI}, the proposed RL controller achieves a better ratio of energy savings to performance loss and a substantial improvement in minimizing the overall ED$^2$P across benchmark applications.

\section{Conclusions}\label{sec:Conclusion}
This paper addresses the critical challenge of power consumption and management within datacenter HPC nodes by presenting a dynamic power adjustment approach that seeks to minimize energy use while preserving application performance. Leveraging offline reinforcement learning, our strategy uses datasets collected from static characterization of HPC nodes across various benchmarks to train an effective RL agent that dynamically optimizes hardware configuration ($PCAP$).

Experimental validations have demonstrated our system's robustness, yielding standard deviations in execution time ranging from a minimum of 0.47\unit{s} to a maximum of 2.91\unit{s} and for energy consumption from a minimum of 0.01\unit{kJ} to a maximum of 1.22 \unit{kJ}, indicating consistent energy savings across both memory-bound and compute-bound benchmark tasks.
Remarkably, the controller proved its efficacy when applied to new and phase-varied applications, achieving an average energy saving of 20.34\% with only a minimal average performance degradation of 7.4\%. Repeated tests have further solidified the controller's capability to significantly enhance energy efficiency. These findings underscore the potential of RL-based control for overcoming the limitations of application- and hardware-centric power management designs. Our research bears particular importance for cloud service providers and HPC cluster administrators aiming to significantly reduce energy expenditure with only a marginal compromise on performance. In the future, we intend to explore the deployment of our algorithms in GPU-intensive tasks and wider cluster job environments. The promising results open new avenues for sustainable computing practices, potentially revolutionizing energy management in datacenters.

While our approach has successfully reduced energy consumption with minimal impact on performance, we have so far limited our use of progress as an application-agnostic signal for online performance to applications that can exhibit an iterative behavior, with a heartbeat that occurs often enough during a control interval and is somewhat stable in rate for a given kernel. Future work will concentrate on widening this approach, from extracting a stable signal for a wider set of metrics (like OpenMP callbacks or MPI activity) and dealing with applications than iterate over a heterogeneous workload. This next stage of research will continue to emphasize balancing execution time constraints with the overarching goal of energy-efficient computing, ensuring that optimal performance is achieved within the required timeframe without compromising on energy savings.

\section*{Acknowledgment}\label{sec:acknowledgment}
Results presented in this paper were obtained by using the Chameleon testbed supported by the National Science Foundation. The dataset and the codes used for the entire experiment, including the static characterization, training, and testing, are hosted on our {GitHub repository}. In compliance with the review guidelines, we have anonymized the GitHub repository details during the evaluation period. We also acknowledge the use of AI tools for language improvements and text generation in certain sections of this paper. All results and generated content were carefully reviewed by the authors for accuracy. Responsibility for the content of this paper lies solely with the authors.

\balance
\bibliographystyle{ACM-Reference-Format}
\bibliography{references}

\end{document}